\documentclass[sigconf]{aamas} 

\usepackage{balance} 

\usepackage{caption}
\usepackage{subcaption}
\usepackage{enumitem}
\usepackage{algorithm}
\usepackage{algorithmic}

\newlength{\subfigwidth}
\newlength{\subfigheight}
\setcopyright{ifaamas}
\acmConference[AAMAS '23]{Proc.\@ of the 22nd International Conference
on Autonomous Agents and Multiagent Systems (AAMAS 2023)}{May 29 -- June 2, 2023}
{London, United Kingdom}{A.~Ricci, W.~Yeoh, N.~Agmon, B.~An (eds.)}
\copyrightyear{2023}
\acmYear{2023}
\acmDOI{}
\acmPrice{}
\acmISBN{}

\acmSubmissionID{591}

\title[AAMAS-2023 Formatting Instructions]{Targeted Search Control in AlphaZero\\for Effective Policy Improvement}

\author{Alexandre Trudeau}
\affiliation{
  \institution{University of Alberta}
  \city{Edmonton}
  \country{Canada}}
\email{trudeau1@ualberta.ca}

\author{Michael Bowling}
\affiliation{
  \institution{University of Alberta}
  \city{Edmonton}
  \country{Canada}}
\email{mbowling@ualberta.ca}

\begin{abstract}
AlphaZero is a self-play reinforcement learning algorithm that achieves superhuman play in chess, shogi, and Go via policy iteration. To be an effective policy improvement operator, AlphaZero’s search requires accurate value estimates for the states appearing in its search tree. AlphaZero trains upon self-play matches beginning from the initial state of a game and only samples actions over the first few moves, limiting its exploration of states deeper in the game tree. We introduce Go-Exploit, a novel search control strategy for AlphaZero. Go-Exploit samples the start state of its self-play trajectories from an archive of \textit{states of interest}. Beginning self-play trajectories from varied starting states enables Go-Exploit to more effectively explore the game tree and to learn a value function that generalizes better. Producing shorter self-play trajectories allows Go-Exploit to train upon more independent value targets, improving value training. Finally, the exploration inherent in Go-Exploit reduces its need for exploratory actions, enabling it to train under more exploitative policies. In the games of Connect Four and 9x9 Go, we show that Go-Exploit learns with a greater sample efficiency than standard AlphaZero, resulting in stronger performance against reference opponents and in head-to-head play. We also compare Go-Exploit to KataGo, a more sample efficient reimplementation of AlphaZero, and demonstrate that Go-Exploit has a more effective search control strategy. Furthermore, Go-Exploit’s sample efficiency improves when KataGo’s other innovations are incorporated.
\end{abstract}


\newcommand{\BibTeX}{\rm B\kern-.05em{\sc i\kern-.025em b}\kern-.08em\TeX}

\begin{document}


\pagestyle{fancy}
\fancyhead{}


\maketitle 


\section{Introduction}\label{sec:Introduction}

AlphaZero \cite{AlphaGo, AlphaGoZero, AlphaZero} is a model-based reinforcement learning (RL) algorithm that has achieved impressive results in two-player, zero-sum games, reaching superhuman play in chess, shogi, and Go. AlphaZero simulates self-play matches with a perfect model of its environment (the rules of the game) to train a neural network that learns a value function and action selection priors over states. Each turn, the value function and priors guide a lookahead search that returns an improved policy. AlphaZero trains its neural network on the self-play matches produced under the improved policies, enabling it to improve its play via policy iteration.

Despite its success, AlphaZero's training suffers from sample inefficiency. In 19x19 Go, AlphaZero requires hundreds of millions of training samples to attain superhuman play (\cite{AlphaZero}, Figure 1c). AlphaZero's sample efficiency depends upon the distribution of states visited and trained upon. Although AlphaZero has a perfect model of its environment, it cannot feasibly visit and learn the optimal value for each state. Instead, AlphaZero trains upon the states that it visits on-policy in simulated self-play matches beginning from the initial state of the game. As in other RL algorithms \cite{Sutton1998}, AlphaZero takes exploratory actions during its self-play matches so that it can train upon a variety of states, enabling it to make more informed action selections in the future. AlphaZero employs simplistic exploration mechanisms during self-play training: randomly perturbing the learned priors guiding search and stochastically selecting actions near the start of self-play matches. As a result, AlphaZero's training procedure exhibits the following limitations:

\begin{enumerate}
    \item Since AlphaZero begins its self-play matches from the initial state of a game, it often transitions into a terminal state before reaching and exploring states deeper in the game tree. In addition, AlphaZero only samples actions over the first few moves of a self-play match, further limiting exploration deeper in the game tree.
    \item AlphaZero's exploration mechanisms cause it to train under weaker, exploratory policies, slowing policy iteration.
    \item AlphaZero only produces a single, noisy value target from a full self-play match, slowing value training.
\end{enumerate}

We hypothesized that AlphaZero could address these limitations, and learn with greater sample efficiency, with a more effective search control strategy. Sutton and Barto define search control as ``the process that selects the starting states and actions for the simulated experiences generated by the model" \cite{Sutton1998}. In AlphaZero, this would amount to strategically choosing the starting state of its simulated trajectories. We propose one such strategy that adheres to four guiding principles. The algorithm should:

\begin{enumerate}[label=(\alph*)]
  \item Continually visit new states throughout the state space to learn their values and a good policy.
  \item Keep track of \textit{states of interest} and have the ability to reliably revisit them for further exploration.
  \item Limit exploration’s bias in the learning targets.
  \item Produce more independent value targets to train upon.
\end{enumerate}

In this paper, we introduce Go-Exploit, a novel search control strategy for AlphaZero. Go-Exploit takes inspiration from Go-Explore \cite{GoExplore} and Exploring Restart Distributions \cite{tavakoli2018exploring}, which begin simulated episodes from previously visited states sampled from a memory. Similarly, Go-Exploit maintains an archive of \textit{states of interest}. At the beginning of a self-play trajectory, the start state is either uniformly sampled from the archive or is set to the initial state of the game. Two factors influencing Go-Exploit’s performance are the definition of \textit{states of interest} and the structure of the archive. In this paper, we experiment with two definitions of \textit{states of interest} and three archive structures.

In the games of Connect Four and 9x9 Go, we show that Go-Exploit exhibits a greater sample efficiency than standard AlphaZero, measured in their average win rates against reference opponents over the course of training and in the results of their head-to-head play. We also compare and contrast Go-Exploit and KataGo \cite{katago}, a more sample efficient reimplementation of AlphaZero. Go-Exploit's search control strategy results in faster learning than KataGo's. Furthermore, Go-Exploit's sample efficiency improves when KataGo's other innovations are incorporated. We conclude by showing how Go-Exploit's adherence to the guiding principles enables it to learn more effectively than AlphaZero.


\section{AlphaZero}

AlphaZero \cite{AlphaZero} represents its learned game knowledge with a neural network $f_{\mathbf{\theta}}(s) = (\mathbf{p}, v)$ parameterized by weights $\mathbf{\theta}$. $f_{\mathbf{\theta}}$ takes a state $s$ as input and outputs a value estimate $v$ estimating the expected game outcome from state $s$ under AlphaZero’s current policy. $f_{\mathbf{\theta}}$ also outputs a vector of action selection priors $\mathbf{p}$ estimating AlphaZero's current policy from state $s$.

AlphaZero improves its play by training upon simulated matches played against itself. These self-play matches begin from the initial state of a game $s_{0}$. On each turn $t$, AlphaZero performs a variant of Monte Carlo Tree Search \cite{MCTS_Survey, Coulom, UCT}, inspired by PUCB \cite{PUCT}, to determine the action $a_t$ that is played. The search tree is initialized with a root node corresponding to $s_{t}$ and a set of edges representing the legal actions. Each edge stores a set of statistics $\{N(s,a),Q(s,a),P(s,a)\}$. $N(s,a)$ is the number of times $(s,a)$ has been traversed during the given search. $Q(s,a)$ is the backed up action-value estimate of $(s,a)$. $P(s,a)$ is the prior probability of selecting action $a$ from state $s$. In each search iteration, the search tree is traversed from the root node using PUCT action selection: $$a = \arg\max_{a} Q(s,a) + c_{\textrm{puct}}P(s,a)\frac{\sqrt{N(s)}}{1 + N(s,a)}$$ $N(s)$ is the number of times state $s$ has been visited during the given search and $c_{\textrm{puct}} > 0$ is an exploration constant. This action selection rule encourages the search to traverse state-action pairs with large action-value estimates $Q(s,a)$, large priors $P(s,a)$, and few search visits $N(s,a)$. Once the search traverses a state-action pair $(s_{L},a)$ with $N(s_{L},a) = 0$, the successor state $s'$ is added as a child of $s_L$ and $f_{\mathbf{\theta}}$ runs inference on the new state: $f_{\mathbf{\theta}}(s') = (\mathbf{p}, v)$. The edge statistics of the legal actions that can be taken from $s'$ are initialized as follows: $\{N(s',a) = 0, Q(s',a) = 0, P(s',a) = p_a\}$, where $p_a$ is the component of $\mathbf{p}$ corresponding to action $a$. Next, the value estimate $v$ is backed up to the state-action pairs that were traversed in the given iteration to update their action-values $Q(s,a)$. $Q(s,a)$ averages the value estimates $v$ of the states in the subtree of $(s,a)$. It estimates the expected outcome from $(s,a)$ based on the value estimates of the likeliest successor states.

Once the final search iteration is complete, the search returns a policy $\mathbf{\pi}_t$. The components of $\mathbf{\pi}_t$ depend upon the distribution of search visits over the root state's actions: $\pi_{t}(a | s_t) = \frac{N(s_t, a)^{1/\tau}}{\sum_{b}N(s_t, b)^{1/\tau}}$, where $\tau > 0$ is a Softmax temperature. In the first $k$ moves of a self-play match, action $a_t$ is sampled from $\mathbf{\pi}_t$. After the first $k$ moves, AlphaZero aims to be more exploitative and plays the action that was most visited during search. When a self-play match reaches a terminal state $s_{T}$ with outcome $z$, AlphaZero produces training samples $(s_{t}, \mathbf{\pi}_{t}, z)$ that are added to an experience replay buffer $B$ \cite{lin1992self, mnih2013playing} with fixed size $|B|$. Once $b_{\textrm{step}}$ new training samples have been added to the replay buffer, $b_{\textrm{batch}}$ training tuples are uniformly sampled to update $f_{\mathbf{\theta}}$. The neural network’s parameters $\mathbf{\theta}$ are updated via stochastic gradient descent on the loss function $$\textrm{loss} = (z - v)^{2} - \mathbf{\pi}_{t}^{T}\log(\mathbf{p}) + c||\mathbf{\theta}||^{2}$$ where $c$ is a regularization constant. Once $f_{\mathbf{\theta}}$ is updated, the next learning step begins.

Training $f_{\mathbf{\theta}}$’s policy head on the policies $\mathbf{\pi}_t$ and the value head on the self-play match outcomes $z$ brings about policy iteration, enabling AlphaZero to learn stronger policies. AlphaZero's search is a policy improvement operator because it concentrates the search visits on the root actions with the largest action-values $Q(s,a)$. This brings about policy improvement as long as the value estimates used in search are sufficiently accurate under the current policy. Then, AlphaZero selects an action $a_t$ with respect to the improved policy $\mathbf{\pi}_t$. Training $f_{\mathbf{\theta}}$’s value head on outcomes produced under the improved policies enables policy evaluation to be with respect to the improved policy. These alternating processes of policy improvement and policy evaluation enable AlphaZero to learn stronger policies over time. However, the scarcity of independent value targets $z$ relative to the policy targets $\mathbf{\pi}_t$ can slow AlphaZero's value training and its subsequent ability to produce improved policies.

\subsection{Exploration in AlphaZero}

The accuracy of $f_{\mathbf{\theta}}$'s value estimates depends upon the distribution of states visited and trained upon. To have accurate value estimates for the diverse set of states that appear during search, AlphaZero must explore the state space during training. AlphaZero ensures exploration by introducing stochasticity into its action selection.

In its search, AlphaZero perturbs the priors over the root node’s actions with noise. When the root node $s_r$ is evaluated by the neural network $f_{\mathbf{\theta}}(s_r) = (\mathbf{p}, v)$, the vector of action probabilities $\mathbf{p}$ is perturbed by a noise vector $\mathbf{d} \sim \textrm{Dir}(\mathbf{\alpha})$ sampled from a Dirichlet distribution. The perturbed priors $P(s_{r},a)$ are computed using the equation $P(s_{r},a) = (1 - \epsilon)p_{a} + \epsilon{d_a}$, where $p_a$ and $d_a$ are components of $\mathbf{p}$ and $\mathbf{d}$, respectively, and $0 < \epsilon < 1$. Randomly perturbing the priors over the root node’s actions causes the policy $\mathbf{\pi}_t$ returned by search to also be perturbed, introducing randomness in AlphaZero’s action selection during self-play.

AlphaZero also achieves exploration through action sampling. Upon the completion of a search, the search visits over the root node’s actions are converted into a policy $\pi_{t}(a | s_t) = \frac{N(s_t, a)^{1/\tau}}{\sum_{b}N(s_t, b)^{1/\tau}}$. The Softmax temperature $\tau$ helps control the level of exploration vs.\@ exploitation in the produced policies. When $\tau = 1.0$, the components of the policy $\pi_{t}$ are directly proportional to the search visits over the root state's actions. When $\tau < 1.0$, the policies produced concentrate a greater portion of the probability on the most visited root actions and are, therefore, more exploitative. When $\tau > 1.0$, the policies produced are more uniform, and thus, more exploratory. For the first $k$ moves of a self-play game, the action that is played is sampled: $a_t \sim \mathbf{\pi}_t$. Sampling actions proportionally to the search visit counts ensures that a variety of actions are tried from a given state, while still favouring the selection of actions that had large action-values $Q(s,a)$ and large priors $P(s,a)$.

The stochasticity in AlphaZero’s action selection presents an exploration-exploitation trade-off. On the one hand, the stochasticity allows AlphaZero to perform policy evaluation at a diverse set of states, improving the accuracy of the value estimates used during search. This enables AlphaZero's search to be a more effective policy improvement operator. On the other hand, the stochasticity causes AlphaZero to generate self-play matches under weaker exploratory policies. This causes policy evaluation to be with respect to the weaker policies and for policy improvement to be with respect to policies $\mathbf{\pi}_t$ perturbed with Dirichlet noise, slowing down policy iteration. AlphaZero manages this exploration-exploitation trade-off with the temperature $\tau$, the number of action sampling moves $k$, and with $\epsilon$, which controls the magnitude of the Dirichlet noise. These hyperparameters must be set large enough to ensure that AlphaZero sufficiently explores the state space, however, they cannot be so large that AlphaZero learns weak policies. This leads action sampling to only take place at the beginning of self-play matches, limiting the exploration of states later in games.


\section{Go-Exploit}

Given the limitations we identified in AlphaZero's training procedure, we adopted the guiding principles in section \ref{sec:Introduction} in designing a new search control strategy for AlphaZero. We took inspiration from Go-Explore \cite{GoExplore} and Exploring Restart Distributions \cite{tavakoli2018exploring} by incorporating an archive of \textit{states of interest} in AlphaZero. Our algorithm, called Go-Exploit, modifies AlphaZero by beginning self-play trajectories from \textit{states of interest} sampled from this archive. This enables Go-Exploit to reliably revisit \textit{states of interest} throughout the game tree (guiding principle (b)) and to complete more self-play trajectories per learning step (guiding principle (d)). Then, the remainder of the self-play trajectory is produced identically to AlphaZero. However, Go-Exploit applies AlphaZero's exploration mechanisms of action sampling and Dirichlet noise from trajectories beginning throughout the game tree, enabling Go-Exploit to effectively explore the state space (guiding principle (a)). Since exploration is built into the ``Go'' step of sampling the start state of a self-play trajectory, we anticipated that Go-Exploit would require less stochasticity in its action selection than AlphaZero, enabling it to learn under more exploitative policies (guiding principle (c)). In this paper, we explore this approach while experimenting with two definitions of \textit{states of interest} and three archive structures to see how they respectively impact the sample efficiency of Go-Exploit.

The way \textit{states of interest} is defined affects the performance of Go-Exploit because it changes the distribution of states that $f_{\mathbf{\theta}}$ is trained upon. \textit{Go-Exploit Visited States} considers nonterminal states visited during self-play games as states of interest because we want action selection to improve from the states visited under AlphaZero's current policy. \textit{Go-Exploit Search States} considers nonterminal search states appearing in trajectories beginning from $s_0$ as states of interest because their value estimates influence the policies $\mathbf{\pi}_t$ returned by search. Each variant of Go-Exploit samples start states from the archive uniformly at random. Since the archive can contain multiple copies of a state, it favours the selection of states that are frequently visited or observed during search.

\subsection{Go-Exploit Visited States}

Go-Exploit Visited States makes simple modifications to AlphaZero. First, it initializes an archive $A$ with the initial state of a game $s_0$. This archive is shared amongst \textit{training actors} that generate self-play trajectories. At the beginning of each self-play trajectory, Go-Exploit uniformly samples a random number $r \in [0,1]$. If $r < \lambda$, Go-Exploit begins its self-play trajectory from $s_0$. If $r \geq \lambda$, Go-Exploit begins its self-play trajectory from a \textit{state of interest} uniformly sampled from the archive. Second, Go-Exploit samples actions from $\pi_t$ for the first $k$ moves of a self-play trajectory
regardless of whether the trajectory begins at $s_0$. Finally, once a self-play trajectory completes, Go-Exploit Visited States adds the nonterminal states that were visited to the archive $A$. We experimented with two variants of Go-Exploit Visited States using two different archives. Go-Exploit Visited States Expanding Archive (GEVE) uses an expanding archive consisting of every visited state. Go-Exploit Visited States Circular Archive (GEVC) employs a fixed-size circular archive consisting of the most recently visited states.

\subsection{Go-Exploit Search States}

Go-Exploit Search States makes similar modifications to AlphaZero. It also employs \textit{training actors} that sample start states from $A$ and produce training data for $f_{\mathbf{\theta}}$, however, they do not add visited states nor search states to the archive. Go-Exploit Search States, instead, concurrently runs \textit{archive actors} responsible for populating the archive. The archive actors always play out complete self-play matches beginning from $s_0$. Once an archive actor's self-play match is complete, it adds all of the nonterminal states that appeared during search into archive $A$. We experimented with two variants of Go-Exploit Search States. Go-Exploit Search States Reservoir Archive (GESR) uses a fixed-size archive and Reservoir Sampling \cite{reservoir} to determine which states are added/removed from the archive. Reservoir Sampling approximates the distribution of states that would be included in the Expanding Archive. The Expanding Archive is not always feasible due to memory constraints. Go-Exploit Search States Circular Archive (GESC) employs a fixed-size circular archive consisting of the most recently observed search states. Pseudocode for each variant of Go-Exploit can be found in Appendix \ref{pseudocode}.

\subsection{Related Work}

Although Go-Exploit is inspired by Go-Explore \cite{GoExplore}, the two algorithms work very differently. In Go-Explore, the ``Go” step is exploitative because it loads a start state associated with a high scoring trajectory. Exploratory actions are taken from this state to discover higher scoring trajectories. In Go-Exploit, on the other hand, the ``Go” step is exploratory because it begins self-play trajectories from states throughout the game tree. Due to the exploration inherent in the sampling of the start states, Go-Exploit can then produce the remainder of its self-play trajectories under more exploitative policies. Hence the name Go-Exploit.

Go-Exploit extends Exploring Restart Distributions (ERDs) \cite{tavakoli2018exploring} to the new setting of AlphaZero. ERDs maintains a \textit{restart memory} of visited states and combines it with the environment's initial state distribution to form the starting state distribution in a simulated environment. Go-Exploit Visited States is analogous to \textit{Uniform Restart}, which uniformly samples the initial state of an episode from the circular restart memory. However, beginning self-play trajectories from previously visited states may not result in the most efficient learning. Go-Exploit Search States extends ERDs beyond visited states. Go-Exploit deliberately uses the notion of \textit{states of interest} when defining which states to include in its archive to allow the inclusion of states that have never been explicitly visited. This enables Go-Exploit Search States to focus its planning updates on successor states appearing in search whose value estimates influence the returned policies.

In MuZero \cite{MuZero}, the successor to AlphaZero that plans with a learned model, greater sample efficiency is also achieved via search control. The authors introduce a variant of MuZero, called \textit{MuZero Reanalyze}, that revisits previously visited states and performs a new search with the latest model. The model is then trained upon the new policy and value targets returned by the search. MuZero Reanalyze and Go-Exploit Visited States are similar in that they both plan from previously visited states. However, MuZero Reanalyze does not simulate new self-play trajectories from these previously visited states, limiting its exploration of the state space.

KataGo \cite{katago} also incorporates search control into the AlphaZero framework. We will describe KataGo's search control procedure in Section \ref{sec:KataGo} and then evaluate its sample efficiency relative to Go-Exploit.

\begin{figure*}[t]
\centering
\begin{tabular}{p{\subfigwidth} p{\subfigwidth}}
\begin{subfigure}[b]{\subfigwidth}
\includegraphics[width=\subfigwidth, height=\subfigheight]{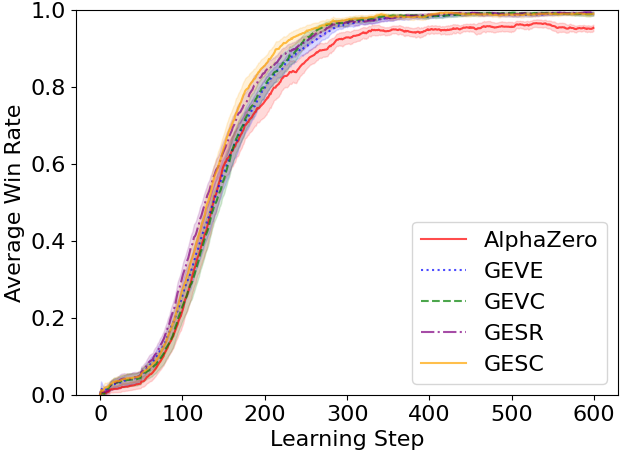}
\caption{Connect Four MCTS-Solver 10x}
\label{Connect_Four_Level_2}
\end{subfigure} &
\begin{subfigure}[b]{\subfigwidth}
\includegraphics[width=\subfigwidth, height=\subfigheight]{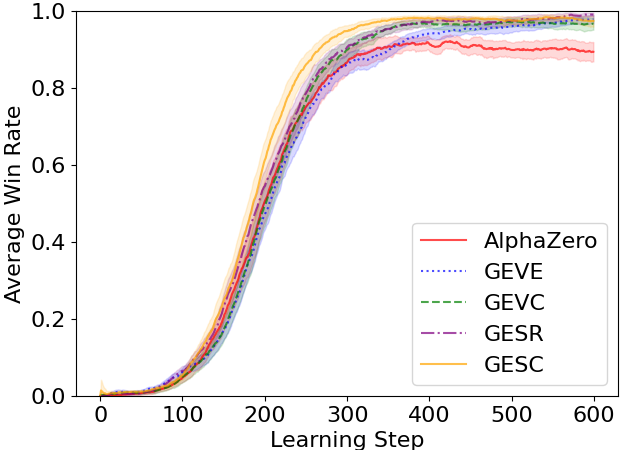}
\caption{Connect Four MCTS-Solver 1000x}
\label{Connect_Four_Level_6}
\end{subfigure} \\

\begin{subfigure}[b]{\subfigwidth}
\includegraphics[width=\subfigwidth, height=\subfigheight]{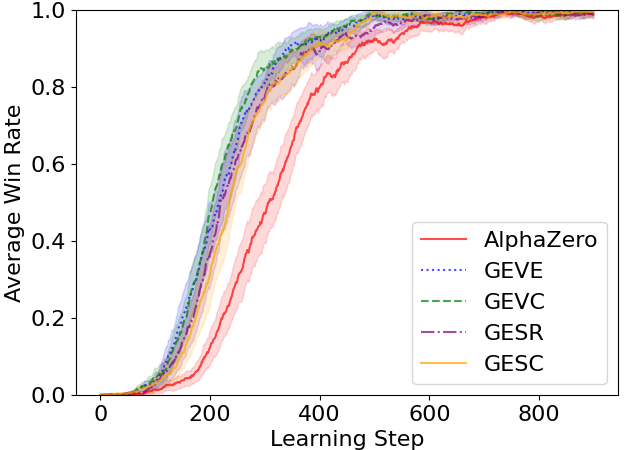}
\caption{9x9 Go MCTS-Solver 10x}
\label{Go_Level_2}
\end{subfigure} &
\begin{subfigure}[b]{\subfigwidth}
\includegraphics[width=\subfigwidth, height=\subfigheight]{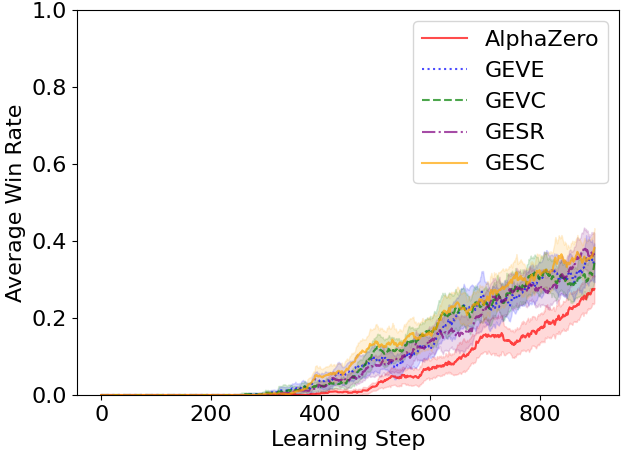}
\caption{9x9 Go MCTS-Solver 1000x}
\label{Go_Level_6}
\end{subfigure}

\end{tabular}
\caption{AlphaZero and Go-Exploit's win rates against MCTS-Solver 10x and 1000x in Connect Four and 9x9 Go. The win rates were averaged over the 30 validation runs and the shaded regions represent $95\%$ confidence intervals.}
\label{Connect_Four_Go_Eval}
\Description{AlphaZero and Go-Exploit's win rates against MCTS-Solver 10x and 1000x in Connect Four and 9x9 Go. The win rates were averaged over the 30 validation runs and the shaded regions represent $95\%$ confidence intervals. In Connect Four, Go-Exploit and AlphaZero exhibit similar learning rates early on in training but ultimately, Go-Exploit achieves a greater asymptotic win rate than AlphaZero. In 9x9 Go, Go-Exploit exhibits a greater learning speed than AlphaZero early on in training but ultimately, they achieve similar asymptotic win rates.}
\end{figure*}


\section{Evaluating Go-Exploit}

Experiments were conducted in Connect Four and 9x9 Go to evaluate the sample efficiencies of the four variants of Go-Exploit relative to AlphaZero and KataGo and to understand how the changes introduced in Go-Exploit impact policy iteration. By performing experiments in Connect Four and 9x9 Go, we evaluated Go-Exploit in two domains with different characteristics and sizes. Connect Four has a smaller state space than 9x9 Go but has a greater percentage of terminal states in its game tree. Go-Exploit and elements of KataGo were coded on top of DeepMind’s OpenSpiel \cite{OpenSpiel} implementation of AlphaZero (our code is publicly available \cite{code}). Experiments were run using OpenSpiel’s versions of Connect Four and Go. Each experiment was run on a node with 48 cores (2 x AMD Milan 7413), 4 NVIDIA A100 gpus, and 498G of memory.

To compare the sample efficiencies of AlphaZero, Go-Exploit, and KataGo, we had to choose a metric measuring sample efficiency. Sample efficiency can be measured by the average performance achieved over a computational budget. We chose this metric because it is less sensitive to training horizon, it accounts for how effectively an algorithm learns throughout training, and it differentiates between algorithms achieving similar asymptotic performance. The average performance over a computational budget can be represented by the ``area under the curve'' (AUC) in a performance vs.\@ learning step graph. In our experiments, training runs lasted a fixed computational budget of 600 learning steps in Connect Four and 900 learning steps in 9x9 Go. An algorithm's performance was measured by its win rate against a fixed reference opponent called MCTS-Solver \cite{MCTSSolver}. Over the course of training, 50 evaluator threads played evaluation matches against different difficulty levels of MCTS-Solver with 1x, 10x, 100x, and 1000x as many search iterations as AlphaZero, Go-Exploit, and KataGo. Equal numbers of evaluation matches were played as player 1 and player 2. Wins, draws, and losses were scored $1$, $0.5$, and $0$, respectively. After each learning step, the win rate against each difficulty level of MCTS-Solver was computed by averaging the evaluation match results over the previous 50 learning steps. In our hyperparameter sweeps, 10 independent training runs were executed for each hyperparameter setting with randomly chosen seeds. Upon their completion, the win rates against each difficulty level of MCTS-Solver were averaged at each learning step. The average win rates were summed over all learning steps to compute the AUC over the computational budget. Ultimately, the AUC achieved against MCTS-Solver 10x was used to select hyperparameter values. To compare the sample efficiencies of AlphaZero, Go-Exploit, and KataGo, an additional 30 validation runs were conducted using the best hyperparameter settings. The AUCs against MCTS-Solver 1x, 10x, 100x, and 1000x were computed to see how the algorithms performed against different fixed reference opponents.

For each algorithm and variant, we swept over the following hyperparameters: the learning rate $lr$ of $f_{\mathbf{\theta}}$, the regularization constant $c$ of the loss function, the Dirichlet distribution parameter $\alpha$, the constant $\epsilon$ affecting the magnitude of the Dirichlet noise, the exploration constant $c_{\textrm{puct}}$ in search, the number of action sampling moves $k$, the probability $\lambda$ of beginning self-play trajectories from $s_0$, the archive type $A_{\textrm{type}}$, the archive size $|A|$, and the Softmax temperature $\tau$. One hyperparameter was swept over at a time. The exact values swept over and the best performing hyperparameter values appear in Appendix \ref{hyperparameter_sweep}. There are additional hyperparameters involved in AlphaZero that do not directly affect the distribution of states visited and trained upon, and thus, are not pertinent to our main investigation. We chose sensible values for our experimental setup that also showed visible learning progress in our domains and held them fixed for all algorithms and variants. This includes the architecture of $f_{\mathbf{\theta}}$, the number of threads, and batch sizes. These values also appear in Appendix \ref{hyperparameter_sweep}.

Once the 30 validation runs were executed, we produced the learning curves appearing in Figure \ref{Connect_Four_Go_Eval} with shaded $95\%$ confidence intervals. In Figures \ref{Connect_Four_Level_2} and \ref{Connect_Four_Level_6}, we observe that in Connect Four, the four variants of Go-Exploit achieve greater AUCs than AlphaZero against MCTS-Solver. Early on in training, AlphaZero and the four variants of Go-Exploit exhibit similar learning rates, but as training progresses, AlphaZero's learning curve levels off to a lower asymptotic win rate. It should be noted that during the hyperparameter sweeps, we observed that AlphaZero could match Go-Exploit's asymptotic win rate with different hyperparameter values but at the cost of a lower AUC (i.e.,\@ AlphaZero attains this asymptotic win rate too slowly). These results suggest that Go-Exploit is able to learn more efficiently than AlphaZero in Connect Four. Figures \ref{Connect_Four_Level_2} and \ref{Connect_Four_Level_6} also illustrate that Go-Exploit achieves even greater AUCs in Connect Four when including search states in its archive rather than visited states. Furthermore, greater sample efficiency is realized when Go-Exploit utilizes a Circular Archive focusing training on the states observed under the most recent policies. In Figures \ref{Go_Level_2} and \ref{Go_Level_6}, we observe that the four variants of Go-Exploit achieve much greater AUCs than AlphaZero in 9x9 Go. Go-Exploit exhibits its superior learning efficiency early on in training with much steeper learning curves than AlphaZero. Ultimately, AlphaZero and the four variants of Go-Exploit reach similar asymptotic win rates. Figure \ref{Go_Level_2} suggests that Go-Exploit may learn marginally faster with visited states rather than search states in the archive. Furthermore, Go-Exploit Visited States obtains a slightly greater AUC with a Circular Archive rather than an Expanding Archive. Comparing the plots in Figure \ref{Connect_Four_Go_Eval} also reveals that Go-Exploit’s gain in sample efficiency is much greater in 9x9 Go than in Connect Four. This suggests that Go-Exploit’s gains in sample efficiency may be greater in larger games. This may be due to the fact that when the search space increases in size, AlphaZero wastes more samples to reach new states deeper in the game tree.

\begin{table}[t]
  \caption{Algorithm 1's win rates in head-to-head matches}
  \label{tab:tournament}
  \begin{tabular}{|l|l|l|l|l|l|}\toprule
  \multicolumn{2}{|c|}{Algorithm} & \multicolumn{2}{c}{Connect Four} & \multicolumn{2}{|c|}{9x9 Go} \\
  \multicolumn{2}{|c|}{} & \multicolumn{2}{c}{Checkpoint} & \multicolumn{2}{|c|}{Checkpoint} \\
  \multicolumn{1}{|c|}{1} & \multicolumn{1}{c}{2} & \multicolumn{1}{|c|}{300} & \multicolumn{1}{|c|}{600} & \multicolumn{1}{c}{300} & \multicolumn{1}{|c|}{900}\\ \midrule
    GEVE & AlphaZero & $0.538$ & $0.643$ & $0.790$ & $0.641$  \\
    GEVC & AlphaZero & $0.483$ & $0.593$ & $0.795$ & $0.655$ \\
    GESR & AlphaZero & $0.513$ & $0.603$ & $0.790$ & $0.670$  \\
    GESC & AlphaZero & $0.582$ & $0.632$ & $0.753$ & $0.652$  \\
    GESC & GEVE & $0.565$ & $0.515$ & $0.471$ & $0.506$  \\
    GESC & GEVC & $0.601$ & $0.530$ & $0.400$ & $0.469$  \\
    GESC & GESR & $0.605$ & $0.519$ & $0.536$ & $0.532$  \\
    GESR & GEVE & $0.505$ & $0.493$ & $0.516$ & $0.502$  \\
    GESR & GEVC & $0.496$ & $0.502$ & $0.436$ & $0.488$  \\
    GEVC & GEVE & $0.483$ & $0.495$ & $0.509$ & $0.488$  \\
    \bottomrule
  \end{tabular}
\end{table}

\begin{figure*}[t!]
\centering
\begin{tabular}{p{\subfigwidth} p{\subfigwidth}}
\begin{subfigure}[b]{\subfigwidth}
\includegraphics[width=\subfigwidth, height=\subfigheight]{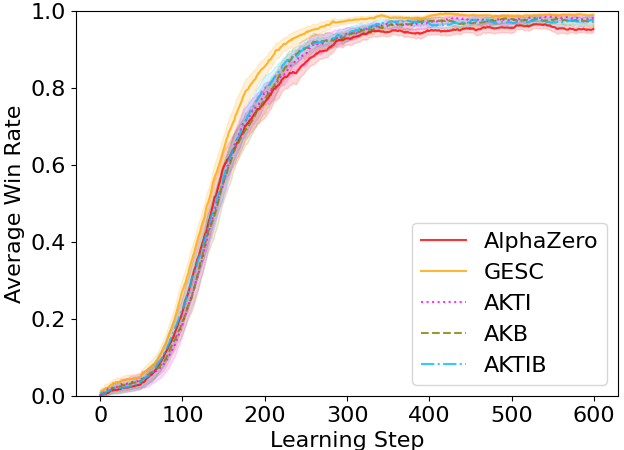}
\caption{AlphaZero with KataGo's search control strategy}
\label{AlphaZero_KataGo_Search_Control}
\end{subfigure} &
\begin{subfigure}[b]{\subfigwidth}
\includegraphics[width=\subfigwidth, height=\subfigheight]{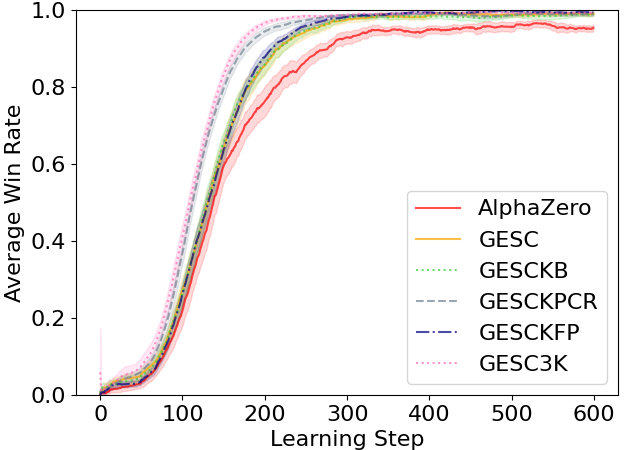}
\caption{Go-Exploit with three of KataGo's modifications}
\label{Go_Exploit_KataGo_Mods}
\end{subfigure}

\end{tabular}
\caption{(a) Comparing the learning speeds of AKTI, AKB, and AKTIB to standard AlphaZero and GESC. (b) Comparing the learning speeds of GESCKB, GESCKPCR, GESCKFP, and GESC3K to standard AlphaZero and GESC. Both plots show the win rates against MCTS-Solver 10x in Connect Four. The shaded regions represent $95\%$ confidence intervals.}
\Description{(a) Comparing the learning speeds of AKTI, AKB, and AKTIB to standard AlphaZero and GESC. AKTI, AKB, and AKTIB exhibit greater learning speeds than AlphaZero but are still inferior to GESC. (b) Comparing the learning speeds of GESCKB, GESCKPCR, GESCKFP, and GESC3K to standard AlphaZero and GESC. GESCKB does not improve upon GESC's learning speed but GESCKPCR, GESCKFP, and GESC3K improve upon GESC's sample efficiency.}
\label{ConnectFourKataGo}
\end{figure*}

To further measure Go-Exploit's learning efficiency relative to AlphaZero, we conducted head-to-head matches between AlphaZero and each variant of Go-Exploit. Head-to-head matches were played using the validation runs' saved neural network checkpoints from learning steps 300 and 600 in Connect Four and learning steps 300 and 900 in 9x9 Go. Each algorithm's 30 neural network checkpoints played the other algorithms' 30 neural network checkpoints in one game as player 1 and one game as player 2. The win rates from the Connect Four and 9x9 Go tournaments appear in Table \ref{tab:tournament}. Each row consists of two algorithms -- Algorithm 1 and Algorithm 2. The remaining entries in the row show Algorithm 1's win rate against Algorithm 2 at the respective checkpoints in Connect Four and 9x9 Go. These results reaffirm what was observed in Figure \ref{Connect_Four_Go_Eval}. In the Connect Four tournament, GESC outperformed AlphaZero and the other variants of Go-Exploit at checkpoint 300, reflecting its superior win rate at learning step 300 in Figures \ref{Connect_Four_Level_2} and \ref{Connect_Four_Level_6}. At checkpoint 600, each variant of Go-Exploit outperformed AlphaZero but none stood out against each other. This mirrors the fact that the four variants of Go-Exploit achieved similar asymptotic win rates that were higher than AlphaZero's in Figures \ref{Connect_Four_Level_2} and \ref{Connect_Four_Level_6}. In the 9x9 Go tournament, each variant of Go-Exploit dominated AlphaZero at checkpoints 300 and 900. This is consistent with Go-Exploit's superior win rate against MCTS-Solver in Figures \ref{Go_Level_2} and \ref{Go_Level_6}. At checkpoint 300, GEVC outperformed both variants of Go-Exploit Search States and marginally beat GEVE, reflecting its superior win rate early on in training in Figure \ref{Go_Level_2}.


\section{Go-Exploit vs. KataGo}\label{sec:KataGo}

KataGo \cite{katago} is an open-source reimplementation of AlphaZero introducing multiple modifications to the original algorithm improving its sample efficiency. In this section, we compare Go-Exploit to the search control procedures introduced in KataGo. Then, we argue that KataGo's other modifications are compatible with Go-Exploit and help it achieve even greater sample efficiency.

\subsection{KataGo's Search Control Strategy}

In the original KataGo paper and its subsequent release notes, key modifications to AlphaZero are highlighted. An additional change, which is not emphasized, is a search control strategy. KataGo's search control procedure involves self-play trajectory initialization and position branching. KataGo initializes self-play trajectories by sampling the first few moves from the policies $\mathbf{p}$ output by $f_{\mathbf{\theta}}$. KataGo occasionally branches trajectories by selecting a different action from the one that was originally played. KataGo also periodically branches from an early position by selecting the action with the greatest value from a set of randomly sampled actions.

To compare the effectiveness of Go-Exploit's search control procedure to KataGo's, we ran OpenSpiel's AlphaZero implementation with KataGo's search control procedure in Connect Four. We ran AlphaZero with KataGo's trajectory initialization (AKTI), with KataGo's branching schemes (AKB), and then with both together (AKTIB). A hyperparameter sweep was conducted for each and then 30 additional validation runs were executed. The learning curves for AlphaZero, AlphaZero with KataGo's search control procedures, and GESC appear in Figure \ref{AlphaZero_KataGo_Search_Control}. This figure illustrates that KataGo's search control strategy achieves a greater AUC than standard AlphaZero, however, it is inferior to the AUC of GESC. KataGo's search control strategy marginally improves AlphaZero's sample efficiency in Connect Four but not nearly as much as Go-Exploit.

\subsection{Go-Exploit's Compatibility With KataGo}

Excluding its trajectory initialization, KataGo's other modifications are compatible with Go-Exploit. KataGo's key modifications alter $f_{\mathbf{\theta}}$'s architecture, the model updates, the feature representation, and search, which are orthogonal to Go-Exploit. To provide evidence that KataGo's other modifications are complementary with Go-Exploit, we incorporated three of KataGo's modifications into GESC to see if they could help it achieve even greater sample efficiency. The first modification was KataGo's branching scheme (GESCKB). While Go-Exploit's trajectory initialization resembles branching, it is not equivalent to KataGo's branching procedure. The second modification was ``Playout Cap Randomization" (GESCKPCR), which randomly varies the number of search iterations performed. The third modification was ``Forced Playouts and Policy Target Pruning" (GESCKFP), which forces visits to certain root actions during search. We ran GESC with each modification individually and then with all three (GESC3K) to see if an even greater sample efficiency could be achieved. For each variant, we performed a hyperparameter sweep and 30 additional validation runs. Their respective learning curves appear in Figure \ref{Go_Exploit_KataGo_Mods}. This figure illustrates that GESC maintains a similar AUC when combined with branching. However, GESC achieves an even greater AUC with Playout Cap Randomization and Forced Playouts + Policy Target Pruning. Furthermore, GESC achieves an even greater AUC when combined with all three. While not definitive, this supports our argument that KataGo's modifications to AlphaZero, other than its trajectory initialization, are complementary with Go-Exploit.


\begin{figure*}[t!]
\centering
\begin{tabular}{p{86mm} p{86mm}}
\begin{subfigure}[b]{86mm}
\includegraphics[width=86mm, height=44mm]{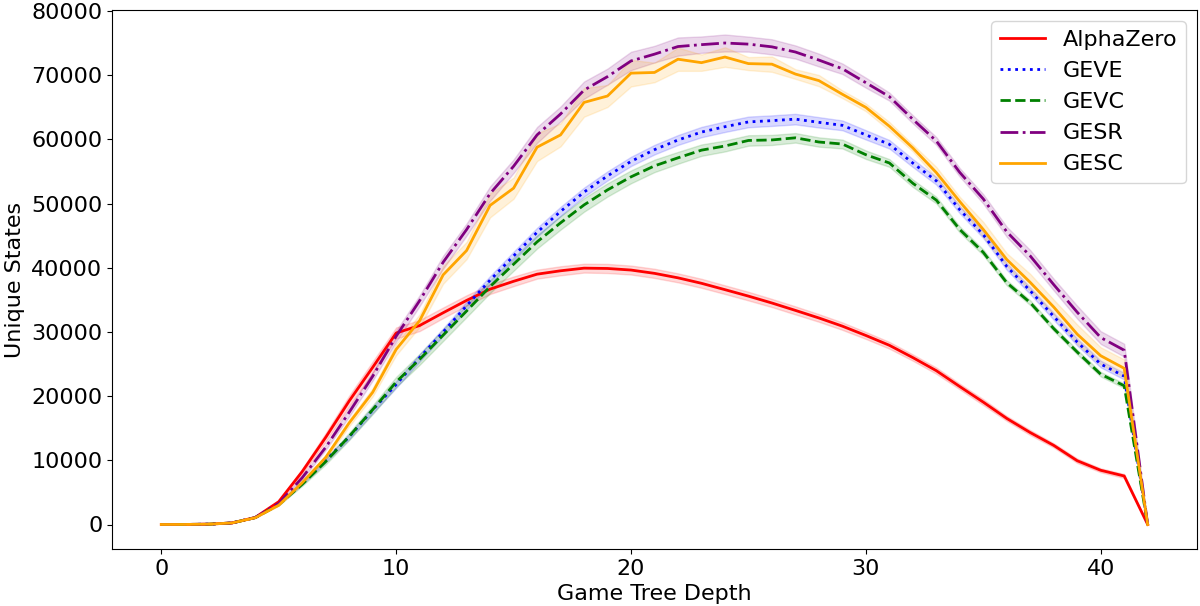}
\caption{Connect Four}
\label{Connect_Four_Unique_States}
\end{subfigure} &
\begin{subfigure}[b]{86mm}
\includegraphics[width=86mm, height=44mm]{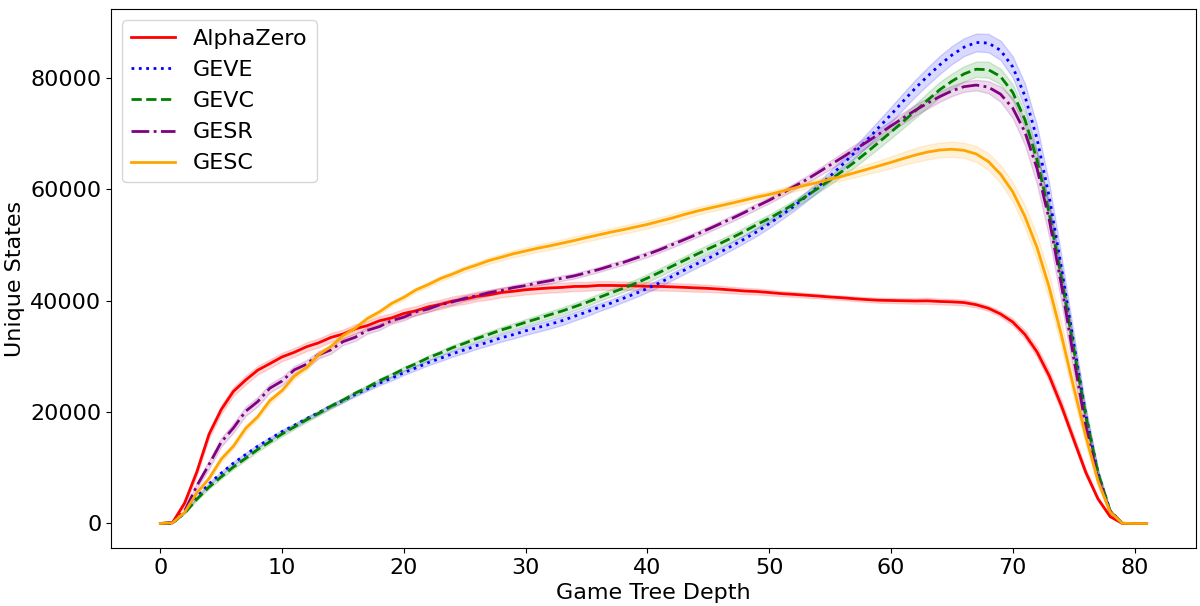}
\caption{9x9 Go}
\label{Go_Unique_States}
\end{subfigure}

\end{tabular}
\caption{The average number of unique nonterminal states visited by AlphaZero and Go-Exploit as a function of game tree depth. The shaded regions represent $95\%$ confidence intervals.}
\Description{The average number of unique nonterminal states visited by AlphaZero and Go-Exploit as a function of game tree depth. In both Connect Four and 9x9 Go, AlphaZero visits a greater number of unique states at earlier game tree depths than Go-Exploit. However, deeper in the game tree, Go-Exploit visits substantially more unique states than AlphaZero.}
\label{Unique_States}
\end{figure*}

\section{Understanding Go-Exploit}

To understand why Go-Exploit learns more efficiently than standard AlphaZero, we collected statistics on the distribution of states visited during self-play and observed during search in the validation runs. In the following subsections, we appeal to these collected statistics and our guiding principles to try to establish why Go-Exploit outperforms AlphaZero in both Connect Four and 9x9 Go.

\subsection{Greater Exploration of the State Space}

We have argued that one of AlphaZero's limitations is that it does not effectively explore states deeper in the game tree. Since AlphaZero always begins its self-play trajectories from $s_0$, it often transitions into a terminal state before reaching and exploring states deep in the game tree. In addition, AlphaZero only samples actions over the first $k$ moves of a self-play game, further limiting exploration deeper in the game tree. These suspicions are confirmed when comparing AlphaZero's distribution of unique visited states to Go-Exploit's. Figures \ref{Connect_Four_Unique_States} and \ref{Go_Unique_States} depict each algorithm's distribution of unique nonterminal states visited by game tree depth in Connect Four and 9x9 Go, respectively. In these plots, we observe that each variant of Go-Exploit visits a greater total number of unique nonterminal states than AlphaZero, particularly deeper in the game tree. At earlier game tree depths, AlphaZero visits more unique states than Go-Exploit. This is expected since AlphaZero begins each self-play trajectory from the initial state of the game whereas Go-Exploit begins its self-play trajectories from states throughout the game tree. In Figure \ref{Connect_Four_Unique_States}, the AlphaZero line gradually decreases at deeper game tree depths because in Connect Four, AlphaZero often transitions into a terminal state and then resets to $s_0$. In Figure \ref{Go_Unique_States}, the AlphaZero line remains fairly level over all game tree depths because there are fewer terminal states in 9x9 Go's game tree. Despite AlphaZero's increased exploration in 9x9 Go, Go-Exploit still outperforms AlphaZero by a greater margin in the larger game of 9x9 Go. It should also be noted that in Figure \ref{Connect_Four_Unique_States}, the slope of the AlphaZero line abruptly decreases at depth 10, which is the number of action sampling moves AlphaZero employs in Connect Four. This confirms the suspicion that only sampling actions over the first $k$ moves of a self-play match hinders AlphaZero's exploration. These plots demonstrate that Go-Exploit is able to more effectively visit and train upon states throughout the state space than AlphaZero (guiding principle (a)) since it begins its self-play trajectories from states throughout the game tree and then subsequently takes exploratory actions from these varied starting points.

Among the variants of Go-Exploit, Go-Exploit Search States visits a greater total number of unique states than Go-Exploit Visited States. In fact, in Connect Four, this is the case over all game tree depths. This may partly explain why Go-Exploit Search States exhibits a greater sample efficiency than Go-Exploit Visited States in Connect Four. On the other hand, in 9x9 Go, Go-Exploit Search States visits a greater number of unique states at earlier game tree depths whereas Go-Exploit Visited States visits a greater number of unique states deeper in the game tree. Furthermore, the percentage difference in visited unique states between Go-Exploit Search States and Go-Exploit Visited States is greater in Connect Four than 9x9 Go. This may partly explain why the differences in sample efficiency between the variants of Go-Exploit are much smaller in 9x9 Go than in Connect Four.

To understand how the differences between AlphaZero and Go-Exploit's state visit distributions impact policy iteration, we compared their value losses over visited states and search states. To establish a fair comparison, we generated 500 self-play matches beginning from the initial state of the game using the final neural network checkpoints from the validation runs. For each visited state, we computed the squared error between the state's value estimate $v_i$ and the outcome of the game $z_i$. For each state observed during search, a trajectory was played to completion without Dirichlet noise and action sampling so that the value loss could also be computed over search states. Table \ref{tab:value_losses} shows each algorithm's average value loss over visited states and search states at checkpoint 600 in Connect Four and checkpoint 900 in 9x9 Go. We observe that each variant of Go-Exploit has a smaller value loss over visited states and search states than AlphaZero.\footnote{It may be surprising that smaller value losses were achieved over search states, however, this is due to there being no added stochasticity in these trajectories.} The fact that Go-Exploit has a smaller value loss over visited states in trajectories beginning from $s_0$ is particularly striking considering that AlphaZero only trains on trajectories beginning from $s_0$ and Go-Exploit does not. Go-Exploit's superior value loss over visited states and search states illustrates that its value function can better predict match outcomes under its current policy and at a greater set of states than AlphaZero. We believe this can be attributed, in part, to Go-Exploit's more effective exploration of the game tree than AlphaZero. The fact that Go-Exploit trains a more accurate and more generalizable value function might be what enables its search to be a more effective policy improvement operator.

\begin{table}[t]
  \caption{Value losses over visited states and search states at checkpoint 600 in Connect Four, checkpoint 900 in 9x9 Go}
  \label{tab:value_losses}
  \begin{tabular}{|l|l|l|l|l|}\toprule
  \multicolumn{1}{|c|}{} & \multicolumn{2}{c}{Connect Four} & \multicolumn{2}{|c|}{9x9 Go} \\
    Algorithm & Visited & Search & Visited & Search \\ \midrule
    AlphaZero & $0.196$ & $0.161$ & $0.293$ & $0.244$ \\
    GEVE & $0.161$ & $0.136$ & $0.227$ & $0.179$ \\
    GEVC & $0.164$ & $0.116$ & $0.223$ & $0.166$ \\
    GESR & $0.170$ & $0.127$ & $0.214$ & $0.165$ \\
    GESC & $0.151$ & $0.108$ & $0.241$ & $0.172$ \\ \bottomrule
  \end{tabular}
\end{table}

\subsection{More Independent Value Targets}

Another potential reason for Go-Exploit's smaller value losses over visited states and search states can be the fact that it produces and trains upon more independent value targets than AlphaZero. In AlphaZero, a new policy target is produced for each visited state whereas only a single independent value target is produced for each self-play trajectory (the outcome of the game). In addition to their scarcity, the value targets trained upon are noisy. The self-play match outcomes are affected by action sampling and Dirichlet noise, and therefore, may not reflect the true values of the visited states. Since Go-Exploit begins its self-play trajectories from states throughout the game tree, its self-play trajectories are shorter, on average, than AlphaZero's. In fact, in Connect Four, AlphaZero completes an average of $147.01$ trajectories per learning step whereas each variant of Go-Exploit completes over $323$, on average. Similarly, in 9x9 Go, AlphaZero completes an average of $74.83$ trajectories per learning step whereas each variant of Go-Exploit completes over $147$, on average. Since Go-Exploit completes more self-play trajectories per learning step than AlphaZero, its experience replay buffer contains more independent value targets, on average, than AlphaZero's. Consistently training on a greater number of independent value targets (guiding principle (d)) may enable Go-Exploit to train a more accurate value function, allowing search to be a more effective policy improvement operator.

\subsection{Training Under More Exploitative Policies}

\begin{table}[t]
  \caption{Hyperparameter values affecting the exploration-exploitation trade-off}
  \label{tab:hyperparameter_values}
  \begin{tabular}{|c|c|c|c|c|c|c|c|c|c|c|}\toprule
  \multicolumn{1}{|c|}{} & \multicolumn{5}{c}{Connect Four} & \multicolumn{5}{|c|}{9x9 Go} \\
     & $c_{\textrm{puct}}$ & $\tau$ & $k$ & $\alpha$ & $\epsilon$ & $c_{\textrm{puct}}$ & $\tau$ & $k$ & $\alpha$ & $\epsilon$ \\ \midrule
    AZ & $1$ & $1$ & $10$ & $1$ & $0.25$ & $1$ & $1$ & $2$ & $0.03$ & $0.1$ \\
    GEVE & $1$ & $1$ & $5$ & $1$ & $0.25$ & $2$ & $1$ & $1$ & $0.03$ & $0.1$ \\
    GEVC & $1$ & $1$ & $10$ & $1$ & $0.1$ & $2$ & $1$ & $1$ & $0.03$ & $0.1$ \\
    GESR & $1$ & $1$ & $2$ & $1$ & $0.25$ & $1$ & $1$ & $2$ & $0.03$ & $0.1$ \\
    GESC & $1$ & $1$ & $10$ & $1$ & $0.25$ & $2$ & $1$ & $1$ & $0.03$ & $0.1$ \\ \bottomrule
  \end{tabular}
\end{table}

In Figures \ref{Connect_Four_Unique_States} and \ref{Go_Unique_States}, we observed that Go-Exploit's search control strategy enables it to more effectively explore the game tree than AlphaZero. Since there is exploration inherent in the sampling of a start state, we hypothesized that Go-Exploit would require less stochasticity in its action selection than AlphaZero (guiding principle (c)). This would allow Go-Exploit to train under more exploitative policies, accelerating its policy iteration. Our hypothesis was mostly confirmed in our hyperparameter sweeps. Table \ref{tab:hyperparameter_values} shows the best performing values for the hyperparameters affecting the exploration-exploitation trade-off for AlphaZero (AZ) and each variant of Go-Exploit in Connect Four and 9x9 Go. In Connect Four, GEVE, GEVC, and GESR are tuned more exploitatively than AlphaZero since they use fewer action sampling moves $k$ or a smaller Dirichlet noise magnitude $\epsilon$. GESC, however, was tuned identically to AlphaZero. In 9x9 Go, GEVE, GEVC, and GESC appear to be tuned more exploitatively than AlphaZero since they use fewer action sampling moves $k$, however, this conclusion is less certain since they employ larger $c_{\textrm{puct}}$ constants, which affect the policies $\mathbf{\pi}_t$ returned by search. Since Go-Exploit is either tuned identically or more exploitatively than AlphaZero in both Connect Four and 9x9 Go, we can conclude that Go-Exploit relies less upon stochastic action selection than AlphaZero to explore the state space. This enables Go-Exploit to produce and train under policies that are inherently more exploitative, accelerating policy iteration.


\section{Conclusion}

In this paper, we have identified limitations in AlphaZero's training procedure and introduced a search control strategy, called Go-Exploit, that mitigates them. In sampling the start states of self-play trajectories from an archive of \textit{states of interest}, Go-Exploit more effectively visits and revisits states throughout the state-space than standard AlphaZero. Furthermore, Go-Exploit produces and trains upon more independent value targets than AlphaZero. These factors enable Go-Exploit to learn a more accurate value function than AlphaZero, allowing search to be a more effective policy improvement operator. In addition, the exploration built into the ``Go'' step reduces Go-Exploit's need for exploratory actions, yielding self-play trajectories produced under stronger, more exploitative policies. These three factors accelerate Go-Exploit's policy iteration, which results in greater sample efficiency. We demonstrated Go-Exploit's ability to learn faster than AlphaZero in Connect Four and 9x9 Go, both measured in its performance against a common reference opponent and in head-to-head matches. We also showed that Go-Exploit utilizes a more effective search control strategy than KataGo and can benefit from KataGo's other improvements.

We have investigated two definitions of \textit{states of interest} and three archive structures but have only sampled from the archive uniformly at random. Future work could investigate new ways of defining \textit{states of interest}, new archive structures, and additional schemes for weighting and/or sampling states in the archive. Additional avenues for future work could include investigating how Go-Exploit can be used with a learned model \cite{MuZero} and in non-deterministic or imperfect information games \cite{Player_Of_Games}.


\bibliographystyle{ACM-Reference-Format} 
\bibliography{sample}


\newpage
\onecolumn
\appendix

\section{Pseudocode}
\label{pseudocode}

\begin{algorithm}[H]
\caption{Go-Exploit}
\label{alg:go_exploit}
\begin{algorithmic}[1] 
\REQUIRE $\textrm{total\_steps}$, $\textrm{num\_training\_actors}$, $\textrm{num\_archive\_actors}$, $A_{\textrm{type}}$, $|A|$, $\textrm{use\_search\_states}$, $|B|$, $b_{\textrm{step}}$, $b_{\textrm{batch}}$, $\lambda$, $\textrm{iters}$, $\alpha$, $\epsilon$, $c_{\textrm{puct}}$, $\tau$, $k$,\newline $lr$, $c$
\STATE Initialize trajectory queue $Q$
\STATE Initialize policy-value network $f_{\theta}$
\STATE $A = \textrm{initialize\_archive}(A_{\textrm{type}}, |A|, s_{0})$
\FOR {$i \in 1,...,\textrm{num\_training\_actors}$}
\STATE $\textrm{training\_actor}(A, \lambda, \textrm{iters}, \alpha, \epsilon, c_{\textrm{puct}}, \tau, k, Q)$
\ENDFOR
\FOR {$i \in 1,...,\textrm{num\_archive\_actors}$}
\STATE $\textrm{archive\_actor}(A, A_{\textrm{type}}, \textrm{use\_search\_states}, \textrm{iters}, \alpha, \epsilon, c_{\textrm{puct}}, \tau, k)$
\ENDFOR
\STATE $\textrm{learner}(\textrm{total\_steps}, |B|, b_{\textrm{step}}, b_{\textrm{batch}}, Q, A, A_{\textrm{type}}, \textrm{use\_search\_states}, lr, c)$
\end{algorithmic}
\end{algorithm}

\begin{algorithm}[H]
\caption{$\textrm{training\_actor}$}
\label{alg:training_actor}
\begin{algorithmic}[1] 
\REQUIRE $A$, $\lambda$, $\textrm{iters}$, $\alpha$, $\epsilon$, $c_{\textrm{puct}}$, $\tau$, $k$, $Q$
\WHILE{True}
\STATE $\textrm{trajectory} = []$
\STATE $t = 0$
\STATE $s_t = \textrm{initialize\_game}()$
\STATE $r = \textrm{rand\_num}(0,1)$
\IF{$r > \lambda$}
\STATE $s_t = A.\textrm{sample}()$
\ENDIF
\WHILE{$s_t$ is not terminal}
\STATE $\pi_t = \textrm{search}(\textrm{iters}, \alpha, \epsilon, c_{\textrm{puct}}, \tau)$
\IF{$t < k$}
\STATE $a_t \sim \pi_{t}$
\ELSE
\STATE $a_t = \arg\max_{a}\pi_{t}[a]$
\ENDIF
\STATE $\textrm{trajectory}.\textrm{add}(s_t, \pi_t)$
\STATE $s_t = \textrm{take\_action}(s_t, a_t)$
\STATE $t = t + 1$
\ENDWHILE
\STATE $z = s_t.\textrm{outcome}()$
\STATE $\textrm{trajectory}.\textrm{set\_outcome}(z)$
\STATE $Q.\textrm{push}(\textrm{trajectory})$
\ENDWHILE
\end{algorithmic}
\end{algorithm}

\begin{algorithm}[H]
\caption{$\textrm{archive\_actor}$}
\label{alg:archive_actor}
\begin{algorithmic}[1] 
\REQUIRE $A$, $A_{\textrm{type}}$, $\textrm{use\_search\_states}$, $\textrm{iters}$, $\alpha$, $\epsilon$, $c_{\textrm{puct}}$, $\tau$, $k$
\WHILE{True}
\STATE $A_{\textrm{temp}} = []$
\STATE $t = 0$
\STATE $s_t = \textrm{initialize\_game}()$
\WHILE{$s_t$ is not terminal}
\STATE $\pi_t$, $\textrm{search\_states} = \textrm{search}(\textrm{iters}, \alpha, \epsilon, c_{\textrm{puct}}, \tau)$
\IF{$t < k$}
\STATE $a_t \sim \pi_{t}$
\ELSE
\STATE $a_t = \arg\max_{a}\pi_{t}[a]$
\ENDIF
\STATE $A_{\textrm{temp}}.\textrm{add}(\textrm{search\_states})$
\STATE $s_t = \textrm{take\_action}(s_t, a_t)$
\STATE $t = t + 1$
\ENDWHILE
\IF{$\textrm{use\_search\_states}$}
\STATE $A.\textrm{update}(A, A_{\textrm{temp}}, A_{\textrm{type}})$
\ENDIF
\ENDWHILE
\end{algorithmic}
\end{algorithm}

\begin{algorithm}[H]
\caption{$\textrm{learner}$}
\label{alg:learner}
\begin{algorithmic}[1] 
\REQUIRE $\textrm{total\_steps}$, $|B|$, $b_{\textrm{step}}$, $b_{\textrm{batch}}$, $Q$, $A$, $A_{\textrm{type}}$, $\textrm{use\_search\_states}$, $lr$, $c$
\STATE $B = \textrm{initialize\_buffer}(|B|)$
\FOR {$\textrm{step} \in 1,...,\textrm{total\_steps}$}
\STATE $A_{\textrm{temp}} = []$
\STATE $\textrm{step\_states} = 0$
\WHILE{$\textrm{step\_states} < b_{\textrm{step}}$}
\STATE $\textrm{trajectory} = Q.\textrm{pop}()$
\FOR{$(s_{t},\boldsymbol{\pi}_{t},z) \in \textrm{trajectory}$}
\STATE $B.\textrm{add}(s_{t}, \boldsymbol{\pi}_{t}, z)$
\STATE $\textrm{step\_states} = \textrm{step\_states} + 1$
\STATE $A_{\textrm{temp}}.\textrm{add}(s_t)$
\ENDFOR
\ENDWHILE
\STATE $b_{\textrm{train}} = B.\textrm{sample}(b_{\textrm{batch}})$
\STATE $f_{\theta}.\textrm{update}(b_{\textrm{train}}, lr, c)$
\IF{NOT use\_search\_states}
\STATE $A.\textrm{update}(A, A_{\textrm{temp}}, A_{\textrm{type}})$
\ENDIF
\ENDFOR
\end{algorithmic}
\end{algorithm}

\begin{algorithm}[H]
\caption{$\textrm{Archive update()}$}
\label{alg:archive_update}
\begin{algorithmic}[1] 
\REQUIRE $A$, $A_{\textrm{temp}}$, $A_{\textrm{type}}$
\FOR{$s \in A_{\textrm{temp}}$}
\IF{$A_{\textrm{type}}$ == ``Expanding"}
\STATE $A$.push($s$)
\ELSIF{$A_{\textrm{type}}$ == ``Circular"}
\IF{$A$.size() $< |A|$}
\STATE $A$.push($s$)
\ELSE
\STATE $A$.pop()
\STATE $A$.push($s$)
\ENDIF
\ELSIF{$A_{\textrm{type}}$ == ``Reservoir"}
\IF{$A$.size() $< |A|$}
\STATE $A$.push($s$)
\ELSE
\STATE $i \sim \textrm{rand\_int}(0, n-1)$
\IF{$i < |A|$}
\STATE $A[i] = s$
\ENDIF
\ENDIF
\STATE $n = n + 1$
\ENDIF
\ENDFOR
\end{algorithmic}
\end{algorithm}

\section{Hyperparameter Sweeps}
\label{hyperparameter_sweep}

\begin{table}[H]
  \centering
  \begin{tabular}{lll}
    \toprule
    & Connect Four & 9x9 Go \\
    \midrule
    Training actors & $700$ & $700$ \\
    Archive actors & $50$ & $50$ \\
    $f_{\mathbf{\theta}}$'s depth & $10$ residual blocks & $10$ residual blocks \\
    $f_{\mathbf{\theta}}$'s width & $256$ filters & $256$ filters \\
    $|B|$ & $2^{17}$ & $2^{17}$ \\
    $b_{step}$ & $2^{12}$ & $2^{12}$ \\
    $b_{batch}$ & $8$ mini-batches of $2^{9}$ & $8$ mini-batches of $2^{9}$ \\
    Search iterations & $100$ & $400$ \\
    Learning steps & $600$ & $900$ \\
    \bottomrule
  \end{tabular}
  \caption[Fixed Hyperparameter Values]{Fixed hyperparameter values}
  \label{Fixed_Params}
\end{table}

\begin{table}[H]
  \centering
  \begin{tabular}{c|c|c|c|c|c}
    \toprule
    & AlphaZero & GEVE & GEVC & GESR & GESC \\
    \midrule
    \multicolumn{1}{c|}{$lr$} & \multicolumn{5}{c}{$\leftarrow[10^{-2}, 10^{-3}, 10^{-4}]\rightarrow$} \\
    \multicolumn{1}{c|}{$c$} & \multicolumn{5}{c}{$\leftarrow[10^{-4}, \mathbf{10^{-5}}, 10^{-6}]\rightarrow$} \\
    \multicolumn{1}{c|}{$\alpha$} & \multicolumn{5}{c}{$\leftarrow[0.03, \mathbf{1.0}, 5.0]\rightarrow$} \\
    \multicolumn{1}{c|}{$\epsilon$} & \multicolumn{5}{c}{$\leftarrow[0.05, 0.1, \mathbf{0.25}, 0.5]\rightarrow$} \\
    \multicolumn{1}{c|}{$c_{puct}$} & \multicolumn{5}{c}{$\leftarrow[0.5, \mathbf{1.0}, 2.0, 4.0]\rightarrow$} \\
    $k$ & $[5, \mathbf{10}, 20, 30]$ & $[2, 5, \mathbf{10}, 20]$ & $[5, \mathbf{10}, 20]$ & $[1, 2, 5, \mathbf{10}, 20]$ & $[5, \mathbf{10}, 20]$ \\
    \multicolumn{1}{c|}{$\lambda$} & \multicolumn{1}{c|}{N/A} & \multicolumn{4}{c}{$\leftarrow[0, 0.01, \mathbf{0.1}, 0.25]\rightarrow$} \\
    \multicolumn{1}{c|}{$|A|$} & \multicolumn{1}{c|}{N/A} & \multicolumn{1}{c|}{N/A} & \multicolumn{3}{c}{$\leftarrow[10^{5}, \mathbf{10^{6}}, 2{\times}10^{6}]\rightarrow$} \\
    \multicolumn{1}{c|}{$\tau$} & \multicolumn{5}{c}{$\leftarrow[0.5, 0.75, \mathbf{1.0}, 2.0]\rightarrow$} \\
    \bottomrule
  \end{tabular}
  \caption[Hyperparameter Values Swept Over in Connect Four]{Hyperparameters swept over in Connect Four. The hyperparameters were swept over in descending order in the table ($lr$ first and $\tau$ last). When a set of hyperparameter values is bounded with arrows ($\leftarrow[\dots]\rightarrow$), it indicates that this set of hyperparameter values was swept over by each algorithm in the column. The values that are bolded were the values that hyperparameters were set to prior to being swept over.}
  \label{Connect_Four_Sweep}
\end{table}

\begin{table}[H]
  \centering
  \begin{tabular}{c|c|c|c|c|c}
    \toprule
    & AlphaZero & GEVE & GEVC & GESR & GESC \\
    \midrule
    $lr$ & $10^{-3}$ & $10^{-3}$ & $10^{-3}$ & $10^{-3}$ & $10^{-3}$ \\
    $c$ & $10^{-5}$ & $10^{-5}$ & $10^{-5}$ & $10^{-5}$ & $10^{-5}$ \\
    $\alpha$ & $1.0$ & $1.0$ & $1.0$ & $1.0$ & $1.0$ \\
    $\epsilon$ & $0.25$ & $0.25$ & $0.1$ & $0.25$ & $0.25$ \\
    $c_{puct}$ & $1.0$ & $1.0$ & $1.0$ & $1.0$ & $1.0$ \\
    $k$ & $10$ & $5$ & $10$ & $2$ & $10$ \\
    $\lambda$ & N/A & $0.1$ & $0.1$ & $0.0$ & $0.01$ \\
    $|A|$ & N/A & N/A & $10^{6}$ & $10^{6}$ & $10^{5}$ \\
    $\tau$ & $1.0$ & $1.0$ & $1.0$ & $1.0$ & $1.0$ \\
    \bottomrule
  \end{tabular}
  \caption[Best Hyperparameter Values in Connect Four]{Best hyperparameter values in Connect Four}
  \label{Connect_Four_Best_Params}
\end{table}

\begin{table}[H]
  \centering
  \begin{tabular}{c|c|c|c|c|c}
    \toprule
    & AlphaZero & GEVE & GEVC & GESR & GESC \\
    \midrule
    \multicolumn{1}{c|}{$lr$} & \multicolumn{5}{c}{$\leftarrow[10^{-2}, 10^{-3}, 10^{-4}]\rightarrow$} \\
    \multicolumn{1}{c|}{$c$} & \multicolumn{5}{c}{$\leftarrow[10^{-4}, \mathbf{10^{-5}}, 10^{-6}]\rightarrow$} \\
    \multicolumn{1}{c|}{$\alpha$} & \multicolumn{5}{c}{$\leftarrow[\mathbf{0.03}, 1.0, 5.0]\rightarrow$} \\
    \multicolumn{1}{c|}{$\epsilon$} & \multicolumn{5}{c}{$\leftarrow[0.05, \mathbf{0.1}, 0.25, 0.5]\rightarrow$} \\
    \multicolumn{1}{c|}{$c_{puct}$} & \multicolumn{5}{c}{$\leftarrow[0.5, 1.0, \mathbf{2.0}, 4.0]\rightarrow$} \\
    \multicolumn{1}{c|}{$k$} & \multicolumn{5}{c}{$\leftarrow[1, 2, \mathbf{5}, 10]\rightarrow$} \\
    \multicolumn{1}{c|}{$\lambda$} & \multicolumn{1}{c|}{N/A} & \multicolumn{4}{c}{$\leftarrow[0.01, \mathbf{0.1}, 0.25]\rightarrow$} \\
    \multicolumn{1}{c|}{$|A|$} & \multicolumn{1}{c|}{N/A} & \multicolumn{1}{c|}{N/A} & \multicolumn{1}{c|}{$[10^{5}, \mathbf{10^{6}}, 2{\times}10^{6}]$} & \multicolumn{2}{c}{$\leftarrow[10^{5}, \mathbf{10^{6}}, 2{\times}10^{6}, 5{\times}10^{6}]\rightarrow$} \\
    \multicolumn{1}{c|}{$\tau$} & \multicolumn{5}{c}{$\leftarrow[0.5, 0.75, \mathbf{1.0}, 2.0]\rightarrow$} \\
    \bottomrule
  \end{tabular}
  \caption[Hyperparameter Values Swept Over in 9x9 Go]{Hyperparameters swept over in 9x9 Go. The hyperparameters were swept over in descending order in the table ($lr$ first and $\tau$ last). When a set of hyperparameter values is bounded with arrows ($\leftarrow[\dots]\rightarrow$), it indicates that this set of hyperparameter values was swept over by each algorithm in the column. The values that are bolded were the values that hyperparameters were set to prior to being swept over.}
  \label{Go_Sweep}
\end{table}

\begin{table}[H]
  \centering
  \begin{tabular}{c|c|c|c|c|c}
    \toprule
    & AlphaZero & GEVE & GEVC & GESR & GESC \\
    \midrule
    $lr$ & $10^{-3}$ & $10^{-3}$ & $10^{-3}$ & $10^{-3}$ & $10^{-3}$ \\
    $c$ & $10^{-5}$ & $10^{-5}$ & $10^{-5}$ & $10^{-5}$ & $10^{-5}$ \\
    $\alpha$ & $0.03$ & $0.03$ & $0.03$ & $0.03$ & $0.03$ \\
    $\epsilon$ & $0.1$ & $0.1$ & $0.1$ & $0.1$ & $0.1$ \\
    $c_{puct}$ & $1.0$ & $2.0$ & $2.0$ & $1.0$ & $2.0$ \\
    $k$ & $2$ & $1$ & $1$ & $2$ & $1$ \\
    $\lambda$ & N/A & $0.1$ & $0.1$ & $0.1$ & $0.1$ \\
    $|A|$ & N/A & N/A & $10^{6}$ & $2\times10^{6}$ & $2\times10^{6}$ \\
    $\tau$ & $1.0$ & $1.0$ & $1.0$ & $1.0$ & $1.0$ \\
    \bottomrule
  \end{tabular}
  \caption[Best Hyperparameter Values in 9x9 Go]{Best hyperparameter values in 9x9 Go}
  \label{Go_Best_Params}
\end{table}

\section{Additional Plots}
\label{additional_plots}

\begin{figure}[H]
\centering
\begin{subfigure}{.5\linewidth}
  \centering
\includegraphics[width=1.0\linewidth]{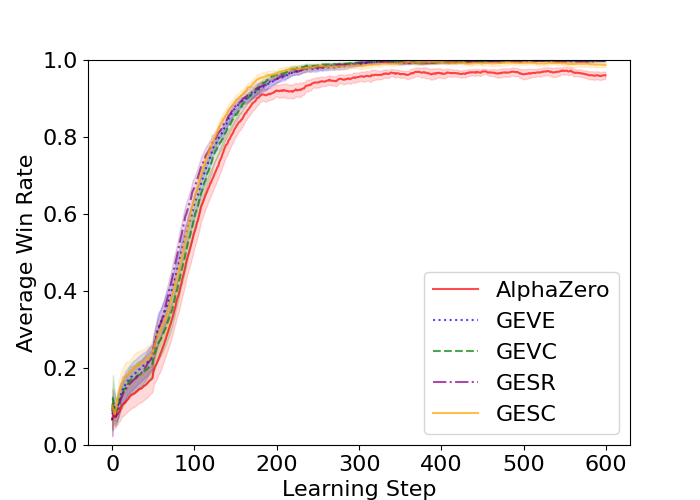}
  \caption{MCTS-Solver 1x}
  \label{Connect_Four_Level_0}
\end{subfigure}%
\begin{subfigure}{.5\linewidth}
  \centering
  \includegraphics[width=1.0\linewidth]{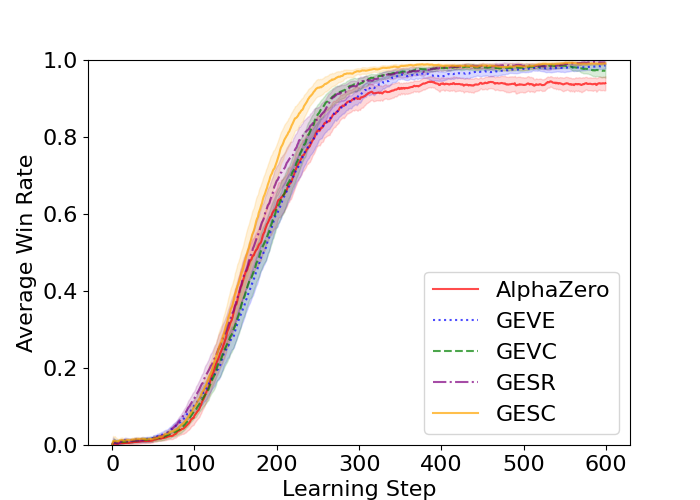}
  \caption{MCTS-Solver 100x}
  \label{Connect_Four_Level_4}
\end{subfigure}%
\caption{AlphaZero and Go-Exploit's win rates against MCTS-Solver 1x and 100x in Connect Four. The win rates were averaged over the 30 validation runs and the shaded regions represent $95\%$ confidence intervals.}.
\label{ExtraConnectFourResults}
\end{figure}

\begin{figure}[H]
\centering
\begin{subfigure}{.5\linewidth}
  \centering
\includegraphics[width=1.0\linewidth]{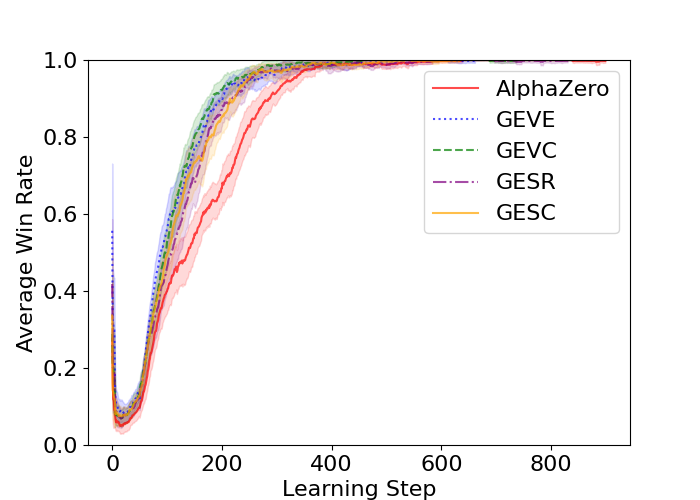}
  \caption{MCTS-Solver 1x}
  \label{Go_Level_0}
\end{subfigure}%
\begin{subfigure}{.5\linewidth}
  \centering
  \includegraphics[width=1.0\linewidth]{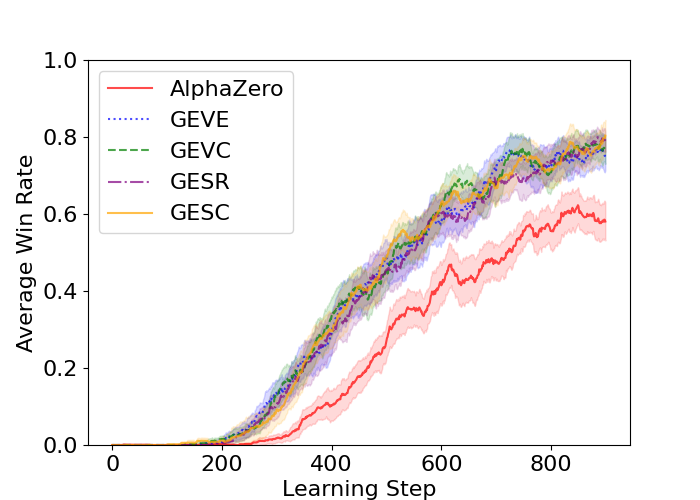}
  \caption{MCTS-Solver 100x}
  \label{Go_Level_4}
\end{subfigure}%
\caption{AlphaZero and Go-Exploit's win rates against MCTS-Solver 1x and 100x in 9x9 Go. The win rates were averaged over the 30 validation runs and the shaded regions represent $95\%$ confidence intervals.}.
\label{ExtraGoResults}
\end{figure}

\end{document}